\pdfoutput=1

\documentclass[11pt]{article}

\usepackage[]{acl}
\usepackage{listings}
\usepackage{times}
\usepackage{latexsym}
\usepackage{multirow}
\usepackage{float}
\usepackage{graphicx}

\usepackage{mathtools}
\usepackage{algorithm,algpseudocode}
 
\newcommand{\eg}{\textit{e.g.}}
\newcommand{\ie}{\textit{i.e.}}

\usepackage[T1]{fontenc}

\usepackage[utf8]{inputenc}
\usepackage{booktabs}
\usepackage{microtype}

\usepackage{amssymb}
\usepackage{amsmath}

\title{Contrastive Learning for Prompt-Based Few-Shot Language Learners}

\author{Yiren Jian \\ Dartmouth College \\
        \texttt{yiren.jian.gr@dartmouth.edu} \\
        \AND
        Chongyang Gao \\ Northwestern University \\ \texttt{cygao@u.northwestern.edu} \\ \And
        Soroush Vosoughi \\ Dartmouth College \\ \texttt{soroush@dartmouth.edu} \\  }

\begin{document}
\maketitle
\begin{abstract}
The impressive performance of GPT-3 using natural language prompts and in-context learning has inspired work on better fine-tuning of moderately-sized models under this paradigm. Following this line of work, we present a contrastive learning framework that clusters inputs from the same class for better generality of models trained with only limited examples. Specifically, we propose a supervised contrastive framework that clusters inputs from the same class under different augmented "views" and repel the ones from different classes. We create different "views" of an example by appending it with different language prompts and contextual demonstrations. Combining a contrastive loss with the standard masked language modeling (MLM) loss in prompt-based few-shot learners, the experimental results show that our method can improve over the state-of-the-art methods in a diverse set of 15 language tasks. Our framework makes minimal assumptions on the task or the base model, and can be applied to many recent methods with little modification. The code will be made available at: https://github.com/yiren-jian/LM-SupCon.
\end{abstract}

\section{Introduction} 
The prompt-based fine-tuning method reduces the gap between pre-training and fine-tuning by forming the fine-tuning task into a masking language problem. A language prompt is a piece of text appended to the query input enabling the model to come up with better predictions \cite{Schick2021ExploitingCF, tam2021improving}. For instance, by feeding a language model with \textit{"The story is not worth reading, a truly \underline{ }\underline{ }\underline{ }\underline{ } one."}, the model assigns a higher probability for the blank to be filled with \textit{"terrible"} than \textit{"great"}. Here, \textit{"a truly \underline{ }\underline{ }\underline{ }\underline{ } one."} is called the template of the prompt and \textit{"terrible"} or \textit{"great"} is the label word. Recently, LM-BFF \cite{gao2021making} shows that appending demonstrations (\eg \textit{"This is an amazing movie, a truly great one"}) to inputs can help the model to better understand the label word, leading to further improved results.

\begin{figure*}[!ht]
\centering
\includegraphics[width=1\textwidth]{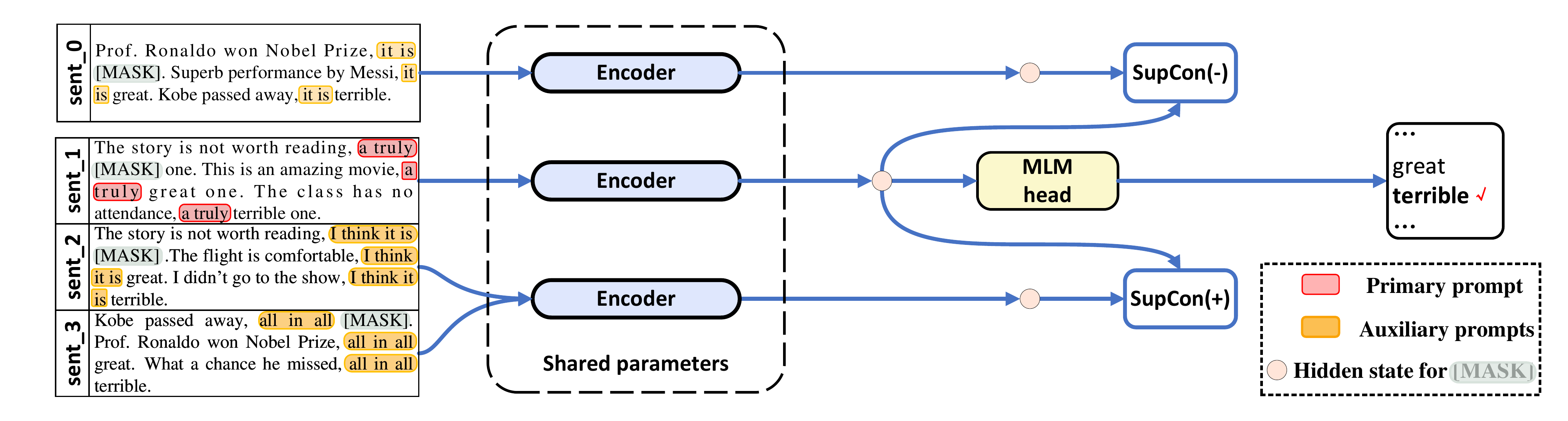}
\caption{Overview of our proposed method. Besides the standard prompt-base MLM loss on label words "great" and "terrible", we introduce a SupCon loss on multi-views of input text. The positive pair is sentences (with sampled templates and/or demonstrations) in the same class, e.g. $\text{sent}_{1}$ and $\text{sent}_{3}$, or itself with a different template and demonstrations, e.g. $\text{sent}_{1}$ and $\text{sent}_{2}$. The negative sentence pair is input sentences (with sampled templates and/or demonstrations) in different classes, e.g. $\text{sent}_{1}$ and $\text{sent}_{0}$.}
\label{fig:overview}
\end{figure*}

In this work, we show that Supervised Contrastive Learning (SupCon) \cite{SupCon} at the feature space can be beneficial during the \emph{fine-tuning} of prompt-based few-shot language learners, with proper data augmentation. 

Data augmentation is the key component of SupCon. While there exists many augmentation techniques like Cutmix \cite{yun2019cutmix}, Mixup \cite{zhang2018mixup} in computer vision and EDA \cite{wei-zou-2019-eda}, AEDA \cite{DBLP:conf/emnlp/KarimiR021} for text, data augmentation remains challenging. 

However, prompt-based few-shot learners with demonstrations actually provide us with a natural way to create multiple "views" (augmentations) of a single example, \ie, for a fixed set of label words, we can sample different templates and different demonstrations to append to the input text (shown in Figure \ref{fig:overview}). This allows us to construct diverse input texts that are consistent and complete. By applying SupCon to cluster the above two example inputs with very different contents but the same label, our method is able to obtain an additional supervision at the feature space which is crucial if we are only given a few labeled examples. 

The main contributions of our paper are:
\begin{itemize}
    \item A Supervised Contrastive Learning framework for prompt-based few-shot learners.
    \item An effective data augmentation method using prompts for contrastive learning with prompt-based learners. 
\end{itemize}

\section{Related Work \& Background}

\textbf{Few-shot Learning} is often tackled by meta learning \cite{li2021semi, bansal2020self, sharaf2020meta, Jian2020TMT, Jian2021MetaPix}, data augmentation \cite{jian2022embed, Jian2022LabelHalluc, arthaud-etal-2021-shot, wei-etal-2021-shot, kumar-etal-2019-closer}. Inspired by the in-context learning of GPT-3, prompt-based fine-tuning \cite{gao2021making, tam2021improving, Schick2021ExploitingCF} recently becomes dominant in NLP. \citet{DBLP:journals/corr/abs-2109-08754} applies contrastive learning in their few-shot semi-supervised intent classification, by using EDA \cite{wei-zou-2019-eda} as augmentation method. Different from \citet{DBLP:journals/corr/abs-2109-08754}, our method applies to prompt-based fine-tuning, and we show in experiments that our proposed augmentation outperforms EDA.

\noindent \textbf{Supervised Contrastive Loss}. SupCon is a special form of contrastive learning \cite{pmlr-v119-chen20j, SimCLRv2, tian2019contrastive, tian2019crd, tian2020makes, liu2021reco, LoCo} that clusters two augmented batches at the class level in the feature space. Let $\tilde{x}_{2k-1}, \tilde{x}_{2k}$ be two augmented views of an input batch $x_{k}$; and ${z}_{2k}, {z}_{2k-1}$ to be the features of $\tilde{x}_{2k-1}, \tilde{x}_{2k}$. Then SupCon loss can be computed as
\begin{align}\label{equation:SupCon}
    \mathcal{L}_{\text{\tiny{SupCon}}}=\text{SupCon}(z_{2k-1}, z_{2k}, y_{k})
\end{align}
where $y_{k}$ is the label for batch $x_{k}$. The details of $\text{SupCon}$ can be found in \citet{SupCon}.

\section{Method}
\textbf{Problem formulation}. Following the few-shot setting in LM-BFF, we assume to have access to a pre-trained language model $\mathcal{M}$, datasets $\mathcal{D}_{\text{train}}$ and $\mathcal{D}_{\text{test}}$ with label space $\mathcal{Y}$. There are only $K=16$ examples per class in $\mathcal{D}_{\text{train}}$.

\textbf{Fine-tuning with prompts and demonstrations}. Prompt-based methods treat a classification problem as a masked language modeling (MLM) problem. They take as input a sentence ($\text{sent}$) and a masked template ($\text{temp}$) (i.e., $x_\text{prompt} = \text{sent}, \text{temp}(\text{[mask]})$), and find the best token to fill in the [mask]. This leads to a MLM loss $\mathcal{L}_{\text{MLM}} = \text{MLM}(x_{\text{prompt}}, y)$, where $y$ is the label word corresponding to $x_{\text{prompt}}$. LM-BFF \cite{gao2021making} further appends demonstrations of label words to improve the results: $x_{\text{prompt+demo}}=\text{sent}_{0}, \text{temp}_{0}(\text{[mask]}),\text{sent}_{i}, \text{temp}_{0}(\text{word}_{i})$
, where $\text{word}_{i}$ is the label word for $\text{sent}_{i}$, and $\text{sent}_{i}$ is sampled from the training set. Then the classification loss becomes:
\begin{align}\label{equation:MLM}
    \mathcal{L}_{\text{MLM}} = \text{MLM}(x_{\text{prompt+demo}}, y)
\end{align}
More mathematical formulation can be found in LM-BFF or our Appendix~\ref{sec:Fpdemo}.

\textbf{Language-based Supervised Contrastive Loss}. For applying SupCon on multi-views of an input text, we need to first obtain two views of a text:
\begin{align*}
    x_{1}=&\text{sent}_{0}, \text{temp}_{0}(\text{[mask]}),\text{sent}_{i}, \text{temp}_{0}(\text{word}_{i}) \\
    x_{2}=&\text{sent}_{0}, \text{temp}_{j}(\text{[mask]}),\text{sent}_{k}, \text{temp}_{j}(\text{word}_{k})
\end{align*}
where $x_{1}$ is identical to $x_{\text{prompt+demo}}$ in LM-BFF. We sample a new template ($\text{temp}_{j}$), demonstration ($\text{sent}_{k}$) and the corresponding label word ($\text{word}_{k}$) to replace those in $x_{1}$, to create a second view of input $x_{2}$. With $x_1$ and $x_2$, we can compute SupCon loss by Equation~\ref{equation:SupCon}. The total loss is then
\begin{align}\label{equation:total-loss}
    \mathcal{L}_{\text{total}} = \mathcal{L}_{\text{MLM}} + \mathcal{L}_{\text{SupCon}}
\end{align}
See our Appendix~\ref{sec:LSCLoss} for more mathematical details.

\textbf{Computational overhead}. We show the algorithm of our method in Algorithm~\ref{alg:main}. In general, our method learns from $\mathcal{L}_{\text{total}} = \mathcal{L}_{\text{MLM}} + \mathcal{L}_{\text{SupCon}}$, whereas baseline LM-BFF learns with $\mathcal{L}_{\text{MLM}}$ only. Learning from $\mathcal{L}_{\text{SupCon}}$ requires one additional forward and backward pass (highlighted in blue in Algorithm~\ref{alg:main}), leading to an increase of computational cost by $\times1.5$.

\begin{figure}[ht]
\centering
\begin{minipage}{1.0\linewidth}
\begin{algorithm}[H]
\caption{Our method}\label{alg:main}
\begin{algorithmic}[1]
\State $Max\_Step = 1000$, \\
$LM$: Language model, \\
$Train\_Set$: Training set, \\ 
$Sample$: Randomly sampling function, \\
$Concatenate$:  The function to concatenate two strings, \\
$CE$:  Cross Entropy loss, \\
$SupCon$:  Supervised Contrastive loss.
\For{i in $Max\_Step$}
    \State $sent, y = Sample(Train\_Set)$
    \State $demo_{1} = Sample(Train\_Set)$
    \State $demo_{2} = Sample(Train\_Set)$
    \State $input_{1} = concatenate(sent, demo_{1})$
    \State $input_{2} = concatenate(sent, demo_{2})$
    
    \textcolor[RGB]{248, 150, 30}{$\triangleright$ Learning from MLM Loss}
    \State \textcolor[RGB]{248, 150, 30}{$output_{1} = LM(input_{1})$}
    \State \textcolor[RGB]{248, 150, 30}{$L_{MLM} = CE(output_{1}, y)$}
    \State \textcolor[RGB]{248, 150, 30}{$L_{MLM}.backward()$}
    \State \textcolor[RGB]{248, 150, 30}{$optimizer.step()$}
    
    \textcolor[RGB]{2, 62, 138}{$\triangleright$ Learning from SupCon Loss}
    \State  \textcolor[RGB]{2, 62, 138}{$output_{2} = LM(input_{2})$}
    \State  \textcolor[RGB]{2, 62, 138}{$L_{SupCon} = SupCon(output_{1}, output_{2})$}
    \State  \textcolor[RGB]{2, 62, 138}{$L_{SupCon}.backward()$}
    \State  \textcolor[RGB]{2, 62, 138}{$optimizer.step()$}
\EndFor
\end{algorithmic}
\end{algorithm}

\end{minipage}
\end{figure}

\section{Experiments}

\begin{table}[!ht]
\begin{center}
\scriptsize

\begin{tabular}{l c c c c }
           \hline
           Task            &  LM-BFF    &  LM-BFF              &  PET       &  PET                \\
                           &            &  + ours              &            &  + ours             \\
           \hline
           SST-2 (acc)     & 89.2 (1.3) &  \textbf{90.6} (0.1) & 88.4 (1.0) & \textbf{89.9} (0.6) \\
           Subj (acc)      & 88.6 (3.3) &  \textbf{90.4} (1.1) & 89.2 (1.5) & \textbf{90.6} (1.6) \\
           SST-5 (acc)     & 47.9 (0.8) &  \textbf{49.5} (1.1) & 46.0 (0.9) & \textbf{48.8} (1.2) \\
           CoLA (Matt.)    & 6.1 (5.3)  &  \textbf{10.2} (5.8) & 3.5 (3.4)  & \textbf{5.9}  (3.3) \\
           TREC (acc)      & 82.8 (3.1) &  \textbf{83.3} (1.5) & 77.8 (9.1) & \textbf{82.3} (4.6) \\
           MNLI (acc)      & 61.0 (2.1) &  \textbf{64.0} (2.0) & 58.2 (1.1) & \textbf{58.9} (3.1) \\
           MNLI-mm (acc)   & 62.5 (2.1) &  \textbf{65.5} (2.7) & 59.8 (1.2) & \textbf{61.0} (3.3) \\
           SNLI (acc)      & 66.9 (2.4) &  \textbf{69.9} (2.4) & 63.1 (2.5) & \textbf{65.7} (3.9) \\
           QNLI (acc)      & 60.7 (1.7) &  \textbf{66.4} (3.5) & 61.5 (3.3) & \textbf{63.5} (3.7) \\
           QQP (acc)       & 62.5 (2.6) &  \textbf{68.8} (3.8) & 61.9 (3.5) & \textbf{65.7} (4.3) \\
           RTE (acc)       & 64.3 (2.7) &  \textbf{65.1} (3.5) & 60.9 (4.7) & \textbf{65.1} (3.5) \\
           MRPC (F1)       & 75.5 (5.2) &  \textbf{78.2} (3.1) & 70.6 (6.0) & \textbf{75.7} (6.1) \\
           MR (acc)        & 83.3 (1.4) &  \textbf{85.8} (0.6) & 85.0 (0.6) & \textbf{85.2} (0.9) \\
           MPQA (acc)      & 83.6 (1.8) &  \textbf{84.6} (1.5) & 81.3 (2.6) & \textbf{81.8} (2.4) \\
           CR (acc)        & 88.9 (1.0) &  \textbf{89.4} (1.0) & 89.3 (1.0) & \textbf{90.5} (0.5)
           \\ \hline

\end{tabular}
\caption{Few-shot experiments of baseline methods and ours. LM-BFF is a prompt-based method with demonstrations of label words and PET is one without demonstrations. The experimental results show the means and standard deviations from 5 different train-test splits.}
\label{table:Main}
\end{center}
\end{table}

\textbf{Evaluation datasets and protocol}. We evaluate our method on 15 classification tasks studied in LM-BFF and follow the same setup as them to allow fair comparisons (see Appendix \ref{sec:BS} for more training details). Contrastive learning algorithms benefit from large batch training. Thus, we report baselines with the same large batch size as ours.

Our method uses a single prompt/template (primary prompt) for the prediction of each task, and a set of prompts (auxiliary prompts) for generating multi-views of inputs for contrastive learning. The primary prompts we used are shown in Appendix~\ref{sec:primary-prompt}. The auxiliary prompts can be either manually designed or generated by a searching algorithm. In this work, we use the top-20 generated prompts from LM-BFF's project page and we randomly sample templates in these 20 prompts to produce second views of our inputs. Unless otherwise noted, we apply \textit{both} random templates and random demonstrations to create second views of inputs for the contrastive learning.

\subsection{Main results on 15 tasks}
We use RoBERTa-base (see Appendix \ref{sec:ExpRoBert} for RoBERTa-large). We compare ours with LM-BFF (a method w/ demonstrations) and PET~\cite{Schick2021ExploitingCF} (a method w/o demonstration). 

Table \ref{table:Main} shows that our SupCon loss can consistently boost the performance of baseline prompt-based fine-tuning method LM-BFF. The introduction of SupCon loss has a maximum improvement of $6.3\%$ in QQP and an average improvement of $2.5\%$ across 15 tasks, likely due to the more generalized representations learned by SupCon. On average, the greater improvements by our model can be seen on the more difficult tasks (see Appendix \ref{sec:diff} for more detail). 

We want to emphasize that the input for baseline LM-BFF already appends different randomly sampled demonstrations at each tuning iteration. Thus, the improvement of our method can not be attributed to the diversity of inputs when learning from $\mathcal{L}_\text{MLM}$ of Equation~\ref{equation:total-loss}, but to the $\mathcal{L}_\text{SupCon}$.

Table \ref{table:Main} also shows that our method works well even for prompt-based methods without demonstrations. PET, which is a method without demonstrations, works consistently worse than LM-BFF. However, with the additional SupCon loss, the few-shot performances of PET can be increased by an average of $2.3\%$. And the gap between having and not having demonstrations can be largely closed (see LM-BFF vs. PET+ours in Table \ref{table:Main}). In some tasks, \eg, SST-2, SST-5, QNLI, QQP, RTE MRPC, MR, and CR, the contribution of our SupCon loss can be even larger than the sole use of the demonstrations for label words.

\subsection{SupCon vs. other losses}

\begin{table}[!ht]
\begin{center}
\tiny
\begin{tabular}{l c c c c c}
           \hline
           Task            &  LM-BFF    &  LM-BFF       &  LM-BFF    &  LM-BFF      &  LM-BFF\\
                           &            &  +Dec         &  +Dec +Lab &  +ConCal        &  +ours\\
           \hline
           SST-2     & 89.2 (1.3) & 90.1 (0.6) & \textbf{90.6} (0.5) & 88.5 (2.0)  &  \textbf{90.6} (0.1) \\
           Subj      & 88.6 (3.3) & 87.3 (3.6) & 88.4 (4.9) & 83.8 (7.3)  &  \textbf{90.4} (1.1) \\
           SST-5     & 47.9 (0.8) & 47.2 (1.0) & 46.5 (0.7) & 47.9 (1.1) &  \textbf{49.5} (1.1) \\
           CoLA      & 6.1 (5.3)  &  9.8 (6.5) &  7.2 (5.2) &  6.7 (4.6) &  \textbf{10.2} (5.8) \\
           TREC      & 82.8 (3.1) & 81.9 (3.0) & 82.3 (3.0) & 71.1 (7.0) &  \textbf{83.3} (1.5) \\
           MNLI      & 61.0 (2.1) & 61.3 (2.1) & 59.4 (1.3) & 61.0 (0.8) &  \textbf{64.0} (2.0) \\
           -mm       & 62.5 (2.1) & 63.2 (2.1) & 61.4 (1.6) & 62.5 (0.8) &  \textbf{65.5} (2.7) \\
           SNLI      & 66.9 (2.4) & 67.0 (3.1) & 65.8 (2.1) & 67.0 (2.9) &  \textbf{69.9} (2.4) \\
           QNLI      & 60.7 (1.7) & 60.0 (2.5) & 60.2 (2.0) & 60.9 (2.0) &  \textbf{66.4} (3.5) \\
           QQP       & 62.5 (2.6) & \textbf{69.0} (1.7) & 65.4 (1.2) & 62.2 (2.7) &  68.8 (3.8) \\
           RTE       & 64.3 (2.7) & \textbf{65.6} (1.5) & 65.3 (2.4) & 60.2 (1.9) &  65.1 (3.5) \\
           MRPC      & 75.5 (5.2) & 69.4 (7.0) & 66.5 (7.0) & \textbf{78.3} (3.1) &  $78.2^{\dagger}$ (3.1) \\
           MR        & 83.3 (1.4) & 85.0 (1.0) & 84.6 (1.2) & 84.0 (1.4) &  \textbf{85.8} (0.6) \\
           MPQA      & 83.6 (1.8) & 82.3 (1.9) & 84.3 (1.4) & 72.3 (13.4) & \textbf{84.6} (1.5) \\
           CR        & 88.9 (1.0) & 89.3 (0.6) & \textbf{89.6} (0.7) & 87.7 (1.1) &  89.4 (1.0) 
           \\ \hline
\end{tabular}
\caption{Comparing our SupCon loss with Decoupling Label Loss (Dec), Label Condition Loss (Lab), and Contextual Calibration (ConCal). $\dagger$ We can achieve stronger performance {\small $80.0 \pm 1.8$} by fixing templates/demonstrations when creating the second view of the input (see Section \ref{sec:SampleTempDemo}).}
\label{table:CompareLoss}
\end{center}
\end{table}

We further show that our method outperforms two latest methods that are designed to improve prompt-based language models. In ADAPET \cite{tam2021improving}, the authors replace the traditional CrossEntropy loss with Decoupling Label Loss and Label Condition Loss in the prompt-based fine-tuning method PET, without demonstrations. Contextual Calibration \cite{poisoning:icml21} calibrates the output probabilities by considering context-free inputs, \ie, " " or "N/A". (Further see Appendix \ref{sec:AdA})

From Table \ref{table:CompareLoss} we observe that on 12 tasks our $\mathcal{L}_\text{SupCon}$ outperforms the other losses, while performs on-par in other tasks. Contextual Calibration does not achieve good results overall. We speculate two reasons for this. First, Contextual Calibration is designed for large models without fine-tuning like GPT  (zero-shot setting). Second, the form of in-context learning in Contextual Calibration is different from the demonstrations we study here.

\subsection{Ensemble vs. our single model}

\begin{table}[!ht]
\begin{center}

\begin{tabular}{l c c}
           \hline
           Task            &  LM-BFF    &  LM-BFF    \\
                           &  +ours     &  ensemble  \\
           \hline
           SST-5 (acc)     &  \textbf{49.5} (1.1) & 48.0 (0.8)\\
           CoLA (Matt.)    &  \textbf{10.2} (5.8) &  7.5 (4,7)\\
           MNLI (acc)      &  \textbf{63.3} (2.4) & 62.2 (1.8)\\
           MNLI-mm (acc)   &  \textbf{65.1} (2.4) & 64.0 (1.8)\\
           QNLI (acc)      &  \textbf{66.4} (3.5) & 63.8 (2.7)\\
           MR (acc)        &  \textbf{85.8} (0.6) & 85.7 (0.7) 
           \\ \hline

\end{tabular}
\caption{Comparing our single model trained with SupCon loss to an ensemble of 20 models. 
}
\label{table:CompareEnsemble}
\end{center}
\end{table}

Our method uses $20$ generated templates (auxiliary prompts) to construct multi-views of input sentences. But only a single prompt (primary prompt) and one set of label words are used for main predictions. Thus, there is only a single model from our method. Here, we compare our model to an ensemble comprised of $20$ models trained separately with the $20$ prompts. From Table \ref{table:CompareEnsemble}, we find that our method even outperforms the ensemble with 20$\times$ more number of parameters, showing that it is a more efficient way to make use of the generated prompts. We speculate that because of the over-fitting nature of few-shot learners, members in the ensemble fail to produce substantial diverse prediction distributions.

\begin{figure}[ht]
\centering
\includegraphics[width=0.52\textwidth]{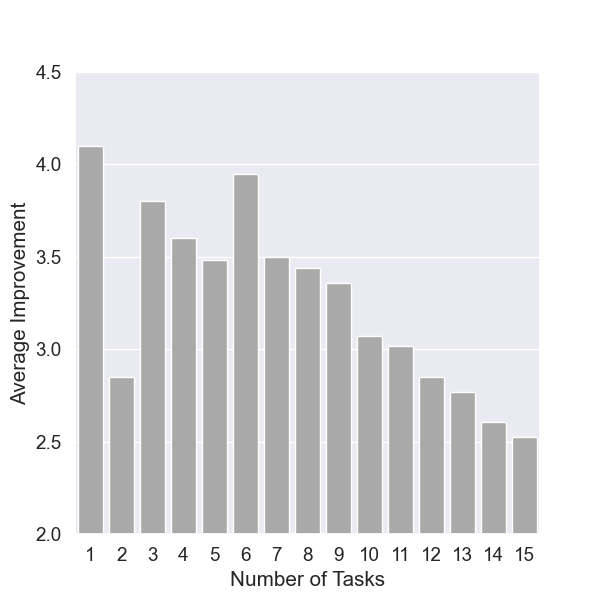}
\caption{The average improvements achieved by our method on the top K hardest tasks, where K goes from 1 to 15.}
\label{fig:bar}
\end{figure}

\section{Improvements vs. Task Difficulty}
\label{sec:diff}
Here, we show that the improvements achieved by our method are greater for tasks with higher difficulty. To show this, we first sort the 15 tasks by base (LM-BFF) performance and use this ranking as a proxy for the difficulty of the task. Next, we report the average improvements achieved by our method on the top K hardest tasks, where K goes from 1 to 15. Figure \ref{fig:bar} shows these results. The first bar corresponds to the improvement achieved by our method on the hardest task, the second bar corresponds to the average improvement achieved by our method on the hardest and second-hardest tasks, and so on. The last bar corresponds to the average improvement on all 15 tasks.

\section{Comparative Experiments}

\subsection{Input augmentation}\label{sec:AblationAug}
The success of contrastive learning heavily relies on the data augmentation. Our method takes advantage of prompt-based language learners and naturally creates multi-views of a single input by appending it with different templates and/or demonstrations. Compared to EDA which includes synonym replacement (SR), random insertion (RI), random swap (RS) and random deletion (RD), our strategy for augmentation does not lead to incomplete and inconsistent sentences, while introducing adequate variations for effective learning.

\begin{table}[ht]
\begin{center}
\scriptsize
\begin{tabular}{l c c c c c c c}
           \hline
           Task & LM-BFF            &     SR    &    RI    &    RS    &  RD   &  EDA    &  ours    \\
           \hline
           SST-2  &  89.2  & 90.7 & \textbf{90.8} & 90.7 & 90.7 & 90.5 &  90.6  \\
           Subj &   88.6  & 90.6 & 90.8 & \textbf{91.0} & 90.5 & 89.1 &  90.4  \\
           SST-5 &  47.9  & 47.7 & 49.2 & 48.2 & 47.9 & 46.7 &  \textbf{49.5}  \\
           CoLA &   6.1   &  5.8 &  6.5 &  4.9 &  4.0  &  3.9 &  \textbf{10.2}  \\
           TREC &   82.8   & 78.1 & 80.7 & 79.0 & 80.7 & 80.6 &  \textbf{83.3}  \\
           MNLI &  61.0    & 61.8 & 62.4 & 61.0 & 58.1 & 58.9 &  \textbf{64.0}  \\
           -mm &   62.5     & 63.6 & 64.8 & 62.7 & 60.3 & 60.9 &  \textbf{65.5}  \\
           SNLI  & 66.9    & 63.1 & 66.4 & 67.2 & 65.2 & 62.2 &  \textbf{69.9}  \\
           QNLI  & 60.7     & 65.3 & 65.3 & \textbf{67.4} & 64.8 & 62.5 &  $66.4^{\dagger}$ \\
           QQP  &   62.5    & 64.5 & 65.8 & 68.0 & 63.2 & 61.0 &  \textbf{68.8}  \\
           RTE  &  64.3     & 61.4 & 61.4 & 61.3 & 62.1 & 61.1 &  \textbf{65.1}  \\
           MRPC &  75.5     & 77.6 & 77.7 & \textbf{79.3} & 78.7 & 79.1 &  $78.2^{\dagger}$  \\
           MR  &  83.3     & 85.5 & 85.5 & 85.5 & 85.3 & 85.6 &  \textbf{85.8}  \\
           MPQA &  83.6     & 82.2 & 84.4 & 84.4  & 83.9 & 82.8 &  \textbf{84.6}  \\
           CR  &  88.9     & 88.9 & 88.2 & 88.3 & 88.5 & 87.1 &  \textbf{89.4}  
           \\ \hline

\end{tabular}
\caption{Comparing our random templates/demonstrations as data augmentation to SR, RI, RS, RD and EDA. Numbers are average of 5 train-test splits.
$\dagger$ We can achieve stronger performance by fixing templates/demonstrations when creating the second view of the input, see Section \ref{sec:SampleTempDemo}.
}
\label{table:CompareEDA10}
\end{center}
\end{table}

The results in Table \ref{table:CompareEDA10} are obtained by applying SR, RI, RS, RD, EDA for $10\%$ of input tokens (Results for $20\%$ are in Appendix \ref{sec:Aug}). In contrast to ours, EDA, etc., for SupCon lead to worse performances than the baseline method in many tasks.

\subsection{Variable templates, demonstrations}\label{sec:SampleTempDemo}
So far, we have shown the results by our method generating multi-views of inputs by appending \textit{both} random templates and demonstrations. However, we find that in some tasks fixed templates with random demonstrations or random templates with fixed demonstration lead to even stronger performances (see Table \ref{table:SampleTempDemo}). For example, sampling demonstrations with fixed templates for MRPC achieves a very strong result ($80.0$), outperforming all other methods in Table \ref{table:CompareEDA10}.

\begin{table}[!ht]
\begin{center}
\scriptsize

\begin{tabular}{l c c c c}
           \hline
           Task            &  LM-BFF    &  - demo    &  + demo    &  + demo      \\
                           &            &  + temp    &  - temp    &  + temp      \\
           \hline
           SST-2 (acc)     & 89.2 (1.3) & \textbf{90.8} (0.3) & 90.5 (0.4) &  90.6 (0.1) \\
           Subj (acc)      & 88.6 (3.3) & \textbf{90.8} (0.8) & 90.6 (1.2) &  90.4 (1.1) \\
           SST-5 (acc)     & 47.9 (0.8) & 49.3 (1.7) & 48.9 (1.8) &  \textbf{49.5} (1.1) \\
           CoLA (Matt.)    & 6.1 (5.3)  &  9.9 (7.5) &  8.5 (5.6) &  \textbf{10.2} (5.8) \\
           TREC (acc)      & 82.8 (3.1) & 83.4 (0.5) & \textbf{86.7} (1.0) &  83.3 (1.5) \\
           MNLI (acc)      & 61.0 (2.1) & 63.4 (3.3) & 63.0 (3.2) &  \textbf{64.0} (2.0) \\
           MNLI-mm (acc)   & 62.5 (2.1) & 65.5 (3.1) & 64.9 (3.4) &  \textbf{65.5} (2.7) \\
           SNLI (acc)      & 66.9 (2.4) & 69.8 (2.4) & 68.5 (1.9) &  \textbf{69.9} (2.4) \\
           QNLI (acc)      & 60.7 (1.7) & 65.4 (3.1) & \textbf{67.0} (3.6) &  66.4 (3.5) \\
           QQP (acc)       & 62.5 (2.6) & \textbf{68.9} (3.2) & 67.8 (1.4) &  68.8 (3.8) \\
           RTE (acc)       & 64.3 (2.7) & 64.9 (3.8) & 62.6 (2.8) &  \textbf{65.1} (3.5) \\
           MRPC (F1)       & 75.5 (5.2) & 79.0 (1.8) & \textbf{80.0} (1.8) &  78.2 (3.1) \\
           MR (acc)        & 83.3 (1.4) & 85.8 (0.7) & 85.4 (0.3) &  \textbf{85.8} (0.6) \\
           MPQA (acc)      & 83.6 (1.8) & 84.0 (1.9) & 84.1 (2.0) &  \textbf{84.6} (1.5) \\
           CR (acc)        & 88.9 (1.0) & 88.6 (0.6) & 88.2 (1.0) &  \textbf{89.4} (1.0) 
           \\ \hline

\end{tabular}
\caption{Different strategies to construct multi-views of input sentences. Fixed demonstrations and sampling templates (- demo + temp), sampling demonstrations and fixed templates (+ demo - temp) and sampling both demonstrations and templates (+ demo + temp). 
}
\label{table:SampleTempDemo}
\end{center}
\end{table}

\section{Limitations}
Since SupCon clusters examples on class level, our framework applies only to classification tasks. Also, our framework requires large GPU memory, as SupCon is an in-batch contrastive loss that needs a large batch size.

\section{Conclusion}
We proposed a novel supervised contrastive learning framework and an effective augmentation method using prompts that can boost the performance of prompt-based language learners and outperform recent work on 15 few-shot tasks.

\section{Ethical Considerations}
As far as we are aware, our proposed work does not have any ethical considerations. However, our work relies on pre-trained language models, which have been shown to be biased in prior work \cite{liang2021towards}. As such, users of such models should be aware of and if possible address such issues. The data and the code for this work will be made available to aid reproducibility.

\bibliography{custom}

\begin{thebibliography}{28}
\expandafter\ifx\csname natexlab\endcsname\relax\def\natexlab#1{#1}\fi

\bibitem[{Arthaud et~al.(2021)Arthaud, Bawden, and
  Birch}]{arthaud-etal-2021-shot}
Farid Arthaud, Rachel Bawden, and Alexandra Birch. 2021.
\newblock \href {https://doi.org/10.18653/v1/2021.eacl-main.90} {Few-shot
  learning through contextual data augmentation}.
\newblock In \emph{Proceedings of the 16th Conference of the European Chapter
  of the Association for Computational Linguistics: Main Volume}, pages
  1049--1062, Online. Association for Computational Linguistics.

\bibitem[{Bansal et~al.(2020)Bansal, Jha, Munkhdalai, and
  McCallum}]{bansal2020self}
Trapit Bansal, Rishikesh Jha, Tsendsuren Munkhdalai, and Andrew McCallum. 2020.
\newblock Self-supervised meta-learning for few-shot natural language
  classification tasks.
\newblock In \emph{Proceedings of the 2020 Conference on Empirical Methods in
  Natural Language Processing (EMNLP)}, pages 522--534.

\bibitem[{Basu et~al.(2021)Basu, lp~Kiun~Chong, Sharaf, Fischer, Rohra, Amoake,
  El{-}Hammamy, Nosakhare, Ramani, and Han}]{DBLP:journals/corr/abs-2109-08754}
Samyadeep Basu, Karine lp~Kiun~Chong, Amr Sharaf, Alex Fischer, Vishal Rohra,
  Michael Amoake, Hazem El{-}Hammamy, Ehi Nosakhare, Vijay Ramani, and Benjamin
  Han. 2021.
\newblock \href {http://arxiv.org/abs/2109.08754} {Semi-supervised few-shot
  intent classification and slot filling}.
\newblock \emph{CoRR}, abs/2109.08754.

\bibitem[{Chen et~al.(2020{\natexlab{a}})Chen, Kornblith, Norouzi, and
  Hinton}]{pmlr-v119-chen20j}
Ting Chen, Simon Kornblith, Mohammad Norouzi, and Geoffrey Hinton.
  2020{\natexlab{a}}.
\newblock \href {https://proceedings.mlr.press/v119/chen20j.html} {A simple
  framework for contrastive learning of visual representations}.
\newblock In \emph{Proceedings of the 37th International Conference on Machine
  Learning}, volume 119 of \emph{Proceedings of Machine Learning Research},
  pages 1597--1607. PMLR.

\bibitem[{Chen et~al.(2020{\natexlab{b}})Chen, Kornblith, Swersky, Norouzi, and
  Hinton}]{SimCLRv2}
Ting Chen, Simon Kornblith, Kevin Swersky, Mohammad Norouzi, and Geoffrey~E
  Hinton. 2020{\natexlab{b}}.
\newblock \href
  {https://proceedings.neurips.cc/paper/2020/file/fcbc95ccdd551da181207c0c1400c655-Paper.pdf}
  {Big self-supervised models are strong semi-supervised learners}.
\newblock In \emph{Advances in Neural Information Processing Systems},
  volume~33, pages 22243--22255. Curran Associates, Inc.

\bibitem[{Gao et~al.(2021)Gao, Fisch, and Chen}]{gao2021making}
Tianyu Gao, Adam Fisch, and Danqi Chen. 2021.
\newblock Making pre-trained language models better few-shot learners.
\newblock In \emph{Association for Computational Linguistics (ACL)}.

\bibitem[{Jian et~al.(2020)Jian, Ahmed, and Torresani}]{Jian2020TMT}
Yiren Jian, Karim Ahmed, and Lorenzo Torresani. 2020.
\newblock Task meta-transfer from limited parallel labels.
\newblock \emph{Meta-Learning workshop, NeurIPS 2020}.

\bibitem[{Jian and Gao(2021)}]{Jian2021MetaPix}
Yiren Jian and Chongyang Gao. 2021.
\newblock \href {https://doi.org/https://doi.org/10.1016/j.imavis.2021.104334}
  {Metapix: Domain transfer for semantic segmentation by meta pixel weighting}.
\newblock \emph{Image and Vision Computing}, 116:104334.

\bibitem[{Jian et~al.(2022)Jian, Gao, and Vosoughi}]{jian2022embed}
Yiren Jian, Chongyang Gao, and Soroush Vosoughi. 2022.
\newblock Embedding hallucination for few-shot language learning.
\newblock In \emph{Proceedings of the 2022 Conference of the North American
  Chapter of the Association for Computational Linguistics: Human Language
  Technologies}. Association for Computational Linguistics.

\bibitem[{Jian and Torresani(2022)}]{Jian2022LabelHalluc}
Yiren Jian and Lorenzo Torresani. 2022.
\newblock Label hallucination for few-shot classification.
\newblock In \emph{Proceedings of the AAAI Conference on Artificial
  Intelligence}.

\bibitem[{Karimi et~al.(2021)Karimi, Rossi, and
  Prati}]{DBLP:conf/emnlp/KarimiR021}
Akbar Karimi, Leonardo Rossi, and Andrea Prati. 2021.
\newblock \href {https://aclanthology.org/2021.findings-emnlp.234} {{AEDA:} an
  easier data augmentation technique for text classification}.
\newblock In \emph{Findings of the Association for Computational Linguistics:
  {EMNLP} 2021, Virtual Event / Punta Cana, Dominican Republic, 16-20 November,
  2021}, pages 2748--2754. Association for Computational Linguistics.

\bibitem[{Khosla et~al.(2020)Khosla, Teterwak, Wang, Sarna, Tian, Isola,
  Maschinot, Liu, and Krishnan}]{SupCon}
Prannay Khosla, Piotr Teterwak, Chen Wang, Aaron Sarna, Yonglong Tian, Phillip
  Isola, Aaron Maschinot, Ce~Liu, and Dilip Krishnan. 2020.
\newblock \href
  {https://proceedings.neurips.cc/paper/2020/file/d89a66c7c80a29b1bdbab0f2a1a94af8-Paper.pdf}
  {Supervised contrastive learning}.
\newblock In \emph{Advances in Neural Information Processing Systems},
  volume~33, pages 18661--18673. Curran Associates, Inc.

\bibitem[{Kumar et~al.(2019)Kumar, Glaude, de~Lichy, and
  Campbell}]{kumar-etal-2019-closer}
Varun Kumar, Hadrien Glaude, Cyprien de~Lichy, and Wlliam Campbell. 2019.
\newblock \href {https://doi.org/10.18653/v1/D19-6101} {A closer look at
  feature space data augmentation for few-shot intent classification}.
\newblock In \emph{Proceedings of the 2nd Workshop on Deep Learning Approaches
  for Low-Resource NLP (DeepLo 2019)}, pages 1--10, Hong Kong, China.
  Association for Computational Linguistics.

\bibitem[{Li and Zhang(2021)}]{li2021semi}
Judith~Yue Li and Jiong Zhang. 2021.
\newblock Semi-supervised meta-learning for cross-domain few-shot intent
  classification.
\newblock \emph{MetaNLP 2021}, page~67.

\bibitem[{Liang et~al.(2021)Liang, Wu, Morency, and
  Salakhutdinov}]{liang2021towards}
Paul~Pu Liang, Chiyu Wu, Louis-Philippe Morency, and Ruslan Salakhutdinov.
  2021.
\newblock Towards understanding and mitigating social biases in language
  models.
\newblock In \emph{International Conference on Machine Learning}, pages
  6565--6576. PMLR.

\bibitem[{Liu et~al.(2021)Liu, Zhi, Johns, and Davison}]{liu2021reco}
Shikun Liu, Shuaifeng Zhi, Edward Johns, and Andrew~J Davison. 2021.
\newblock Bootstrapping semantic segmentation with regional contrast.
\newblock \emph{arXiv preprint arXiv:2104.04465}.

\bibitem[{Schick and Sch{\"u}tze(2021)}]{Schick2021ExploitingCF}
Timo Schick and Hinrich Sch{\"u}tze. 2021.
\newblock Exploiting cloze-questions for few-shot text classification and
  natural language inference.
\newblock In \emph{EACL}.

\bibitem[{Sharaf et~al.(2020)Sharaf, Hassan, and
  Daum{\'e}~III}]{sharaf2020meta}
Amr Sharaf, Hany Hassan, and Hal Daum{\'e}~III. 2020.
\newblock Meta-learning for few-shot nmt adaptation.
\newblock \emph{ACL}, page~43.

\bibitem[{Tam et~al.(2021)Tam, Menon, Bansal, Srivastava, and
  Raffel}]{tam2021improving}
Derek Tam, Rakesh~R Menon, Mohit Bansal, Shashank Srivastava, and Colin Raffel.
  2021.
\newblock Improving and simplifying pattern exploiting training.
\newblock In \emph{Empirical Methods in Natural Language Processing (EMNLP)}.

\bibitem[{Tian et~al.(2020{\natexlab{a}})Tian, Krishnan, and
  Isola}]{tian2019contrastive}
Yonglong Tian, Dilip Krishnan, and Phillip Isola. 2020{\natexlab{a}}.
\newblock Contrastive multiview coding.
\newblock In \emph{Computer Vision -- ECCV 2020}, pages 776--794, Cham.
  Springer International Publishing.

\bibitem[{Tian et~al.(2020{\natexlab{b}})Tian, Krishnan, and
  Isola}]{tian2019crd}
Yonglong Tian, Dilip Krishnan, and Phillip Isola. 2020{\natexlab{b}}.
\newblock Contrastive representation distillation.
\newblock In \emph{International Conference on Learning Representations}.

\bibitem[{Tian et~al.(2020{\natexlab{c}})Tian, Sun, Poole, Krishnan, Schmid,
  and Isola}]{tian2020makes}
Yonglong Tian, Chen Sun, Ben Poole, Dilip Krishnan, Cordelia Schmid, and
  Phillip Isola. 2020{\natexlab{c}}.
\newblock What makes for good views for contrastive learning.
\newblock In \emph{NeurIPS}.

\bibitem[{Wei et~al.(2021)Wei, Huang, Vosoughi, Cheng, and
  Xu}]{wei-etal-2021-shot}
Jason Wei, Chengyu Huang, Soroush Vosoughi, Yu~Cheng, and Shiqi Xu. 2021.
\newblock \href {https://doi.org/10.18653/v1/2021.naacl-main.434} {Few-shot
  text classification with triplet networks, data augmentation, and curriculum
  learning}.
\newblock In \emph{Proceedings of the 2021 Conference of the North American
  Chapter of the Association for Computational Linguistics: Human Language
  Technologies}, pages 5493--5500, Online. Association for Computational
  Linguistics.

\bibitem[{Wei and Zou(2019)}]{wei-zou-2019-eda}
Jason Wei and Kai Zou. 2019.
\newblock \href {https://www.aclweb.org/anthology/D19-1670} {{EDA}: Easy data
  augmentation techniques for boosting performance on text classification
  tasks}.
\newblock In \emph{Proceedings of the 2019 Conference on Empirical Methods in
  Natural Language Processing and the 9th International Joint Conference on
  Natural Language Processing (EMNLP-IJCNLP)}, pages 6383--6389, Hong Kong,
  China. Association for Computational Linguistics.

\bibitem[{Xiong et~al.(2020)Xiong, Ren, and Urtasun}]{LoCo}
Yuwen Xiong, Mengye Ren, and Raquel Urtasun. 2020.
\newblock \href
  {https://proceedings.neurips.cc/paper/2020/file/7fa215c9efebb3811a7ef58409907899-Paper.pdf}
  {Loco: Local contrastive representation learning}.
\newblock In \emph{Advances in Neural Information Processing Systems},
  volume~33, pages 11142--11153. Curran Associates, Inc.

\bibitem[{Yun et~al.(2019)Yun, Han, Oh, Chun, Choe, and Yoo}]{yun2019cutmix}
Sangdoo Yun, Dongyoon Han, Seong~Joon Oh, Sanghyuk Chun, Junsuk Choe, and
  Youngjoon Yoo. 2019.
\newblock Cutmix: Regularization strategy to train strong classifiers with
  localizable features.
\newblock In \emph{Proceedings of the IEEE/CVF International Conference on
  Computer Vision}, pages 6023--6032.

\bibitem[{Zhang et~al.(2018)Zhang, Cisse, Dauphin, and
  Lopez-Paz}]{zhang2018mixup}
Hongyi Zhang, Moustapha Cisse, Yann~N Dauphin, and David Lopez-Paz. 2018.
\newblock mixup: Beyond empirical risk minimization.
\newblock In \emph{International Conference on Learning Representations}.

\bibitem[{Zhao et~al.(2021)Zhao, Wallace, Feng, Klein, and
  Singh}]{poisoning:icml21}
Tony~Z. Zhao, Eric Wallace, Shi Feng, Dan Klein, and Sameer Singh. 2021.
\newblock {Calibrate Before Use: Improving Few-shot Performance of Language
  Models}.
\newblock In \emph{International Conference on Machine Learning (ICML)}.

\end{thebibliography}
\bibliographystyle{acl_natbib}

\clearpage
\appendix

\section{Batch size and learning details}
\label{sec:BS}
We use the same learning rate of $1e^{-5}$ for MLM loss as LM-BFF. To take full advantage of SupCon, we apply large batch sizes (16, 32, 40). We show the batch size and learning rate for SupCon in Table \ref{table:HyperParams}. Note that for results of LM-BFF shown in the main paper, we use the same large batch size of our method to allow for fair comparisons.

We set the batch size to be dividable by the total number of examples in the task and small enough to fit into the GPU memory. The experiments with RoBERTa-base are carried out on one NVIDIA RTX-A6000 with 48 GB of memory. Experiments with RoBERTa-large require 4x NVIDIA RTX-8000 (or RTX-A6000) with 192 (4x 48) GB of momery.

Following LM-BFF, our fine-tuning runs a maximum of 1000 steps.
\setcounter{table}{0}
\renewcommand\thetable{\Alph{section}.\arabic{table}}

\begin{table}[ht]
\begin{center}

\begin{tabular}{l c c}
           \hline
           Task            &  Batch    &  LR   \\
           \hline
           SST-2     & 16 & $1e^{-6}$ \\
           Subj      & 16 & $1e^{-5}$ \\
           SST-5     & 40 & $1e^{-5}$ \\
           CoLA      & 16 & $1e^{-5}$ \\
           TREC      & 32 & $1e^{-5}$ \\
           MNLI      & 24 & $1e^{-5}$ \\
           MNLI-mm   & 24 & $1e^{-5}$ \\
           SNLI      & 32 & $1e^{-5}$ \\
           QNLI      & 16 & $1e^{-5}$ \\
           QQP       & 32 & $1e^{-5}$ \\
           RTE       & 32 & $1e^{-6}$ \\
           MRPC      & 16 & $1e^{-5}$ \\
           MR        & 16 & $1e^{-6}$ \\
           MPQA      & 16 & $1e^{-5}$ \\
           CR        & 32 & $1e^{-5}$ 
           \\ \hline

\end{tabular}
\caption{Batch size and learning rate (LR) for SupCon loss used for each task.}
\label{table:HyperParams}
\end{center}
\end{table}

\section{Fine-tuning with prompts and demonstrations}
\label{sec:Fpdemo}
We also consider LM-BFF as our baseline method due to its state-of-the-art performance in a wide range of few-shot tasks. The given masked language model $\mathcal{M}$ first encodes the input sentence $x_{\text{in}}$ into a sequence of tokens $\tilde{x}_{\text{in}}$ and maps $\tilde{x}_{\text{in}}$ to a sequence of hidden states $\{ \textbf{h}_{1}, \textbf{h}_{2}, ... \textbf{h}_{L} \}$, where $L$ is the length of the sequence and $\textbf{h} \in \mathbb{R}^{d}$, where $d$ is the dimension of the hidden states. For example, in prompt-base fine-tuning, for single sentence text $x_{\text{in}}$ \eg, \textit{"The story is not worth reading."}), the input with the prompt (\eg, \textit{"\text{a truly} \texttt{[MASK]} \text{ one }."}) takes the form of 
\begin{align*}
    x_{\text{prompt}} &= \texttt{[CLS]} x_{\text{in}} \text{, a truly} \texttt{[MASK]} \text{one}. \texttt{[SEP]} \\
                      &\equiv \mathcal{T}(x_{\text{in}}) 
\end{align*}
Then, the model decides whether it is more likely to put the label word \textit{"great"} or \textit{"terrible"} at the $\texttt{[MASK]}$ position. Fine-tuning with this fill-in-the-blank framework has been shown to be superior to standard fine-tuning \cite{Schick2021ExploitingCF}. By mapping the label space $\mathcal{Y}$ to the label words where $\mathcal{V}(y)$ denotes the label word for class $y$, the prediction of the model $\mathcal{M}$ for class $y \in \mathcal{Y}$ can be written as
\begin{align}
    p(y|x_{\text{in}}) &= p(\texttt{[MASK]}=\mathcal{V}(y)|x_{\text{prompt}})\\
                &= \frac{\exp (\textbf{w}_{\mathcal{V}(y)}\cdot \textbf{h}_{\texttt{[MASK]}})} {\sum_{y' \in \mathcal{Y}} \exp (\textbf{w}_{\mathcal{V}(y')}\cdot \textbf{h}_{\texttt{[MASK]}})} 
\end{align}
where $\textbf{w}$ is the weight vector of MLM head.

In LM-BFF, the authors further append demonstrations to the input $x_{prompt}$ to help the model better understand what is \textit{"great"} and \textit{"terrible"}. Formally, $x_{\text{prompt}} \equiv \mathcal{T}(x_{\text{in}})$ and $\tilde{\mathcal{T}}(x_{\text{in}}^{c}, y^{c})$ denote $\mathcal{T}(x_{\text{in}}^{c})$ with $\texttt{[MASK]}$ replaced by the label word $\mathcal{V}(y^{c})$. Then, the input to LM-BFF takes the form of
\begin{align}\label{equation:view0}
    \mathcal{T}(x_{\text{in}}) \oplus \tilde{\mathcal{T}}(x_{\text{in}}^{1}, y^{1}) \oplus ... \oplus \tilde{\mathcal{T}}(x_{\text{in}}^{|\mathcal{Y}|}, y^{|\mathcal{Y}|}) 
\end{align}
In this paper, we use random sampling for the demonstrations, \ie, $x_{\text{in}}^{c}$ is randomly chosen from the training set. The masked language modeling loss is then
\begin{align}
    \mathcal{L}_{\text{MLM}} = \sum_{(x_{\text{in}},y) \in \mathcal{D}_{\text{train}}} -\log p(y|x_{\text{in}})
\end{align}

\setcounter{table}{0}
\renewcommand\thetable{D.\arabic{table}}
\begin{table*}[ht]
\begin{center}
\small
 
\begin{tabular}{l c c}
          \hline
          Task            &  Template    &  Label words      \\
          \hline
          SST-2           & \texttt{<}$S_{1}$\texttt{>} It was \texttt{[MASK]} . & positive: great, negative: terrible \\
          SST-5    & \texttt{<}$S_{1}$\texttt{>} It was \texttt{[MASK]} .  &  v.positive: great, positive: good, neutral: okay, negative: bad, v.negative: terrible \\
          MR       & \texttt{<}$S_{1}$\texttt{>} It was \texttt{[MASK]} . & positive: great, negative: terrible \\
          CR      & \texttt{<}$S_{1}$\texttt{>} It was \texttt{[MASK]} . &  positive: great, negative: terrible \\
          MPQA      & \texttt{<}$S_{1}$\texttt{>} It was \texttt{[MASK]} . &  positive: great, negative: terrible \\
          Subj      & \texttt{<}$S_{1}$\texttt{>} This is \texttt{[MASK]} . &  subjective: subjective, objective: objective \\
          TREC      & \texttt{[MASK]} : \texttt{<}$S_{1}$\texttt{>} &  abbreviation: Expression, entity: Entity, description: Description \\
                 & & human: Human, location: Location, numeric: Number\\
          CoLA      & \texttt{<}$S_{1}$\texttt{>} This is \texttt{[MASK]} . &  grammatical: correct, not\_grammatical: incorrect \\
            \hline
          MNLI       & \texttt{<}$S_{1}$\texttt{>} ? \texttt{[MASK]} ,  \texttt{<}$S_{2}$\texttt{>} &  entailment: Yes, netural: Maybe, contradiction: No \\
          SNLI       & \texttt{<}$S_{1}$\texttt{>} ? \texttt{[MASK]} ,  \texttt{<}$S_{2}$\texttt{>} &  entailment: Yes, netural: Maybe, contradiction: No \\
          QNLI       & \texttt{<}$S_{1}$\texttt{>} ? \texttt{[MASK]} ,  \texttt{<}$S_{2}$\texttt{>} &  entailment: Yes,  not\_entailment: No  \\
          RTE        & \texttt{<}$S_{1}$\texttt{>} ? \texttt{[MASK]} ,  \texttt{<}$S_{2}$\texttt{>} &entailment: Yes,  not\_entailment: No  \\
          MRPC      & \texttt{<}$S_{1}$\texttt{>} \texttt{[MASK]} ,  \texttt{<}$S_{2}$\texttt{>} &  entailment: Yes,  not\_entailment: No  \\
          QQP        & \texttt{<}$S_{1}$\texttt{>} \texttt{[MASK]} ,  \texttt{<}$S_{2}$\texttt{>} & entailment: Yes,  not\_entailment: No\\  \hline

\end{tabular}
\caption{Primary templates and label words used in our experiments.}
\label{table:ManualPrompts_append}
\end{center}
\end{table*}

\section{Language-based Supervised Contrastive Loss}
\label{sec:LSCLoss}
Our method extends the loss $\mathcal{L}_{\text{MLM}}$ with an additional Supervised Contrastive Loss (SupCon). For applying SupCon on multi-views of an input text, we need to first obtain a second view of a text:
\begin{align}
    \tilde{x}_{2k} = Aug(\tilde{x}_{2k-1})
\end{align}
As we show in ablations, traditional data augmentation for text does not work well in the contrastive framework. Thus, we propose obtaining a second view by randomly changing the templates and/or demonstrations:
\begin{align}\label{equation:view1}
    \tilde{x}_{2k-1} = \mathcal{T}_{t_{0}}(x_{\text{in}}) \oplus... \oplus \tilde{\mathcal{T}}_{t_{0}}(x_{\text{in}}^{c}, y^{c}) \oplus ...
\end{align}
\begin{align}\label{equation:view2}
    \tilde{x}_{2k} = \mathcal{T}_{t_{j}}(x_{\text{in}}) \oplus... \oplus \tilde{\mathcal{T}}_{t_{j}}(\hat{x}_{\text{in}}^{c}, y^{c}) \oplus ...
\end{align}
where $T$ denotes a set of pre-defined templates, $t_{j} \in T$ and $t_{j} \neq t_{0}$. $\hat{x}_{\text{in}}^{c}$ is another randomly sampled example as the demonstration text and $\hat{x}_{\text{in}}^{c} \neq x_{\text{in}}$. This strategy serves as a perfect form of augmentation for our purpose as it does not generate incomplete or inconsistent sentences, and since we do not edit the main input, the label for that input stays the same. Furthermore, $\tilde{x}_{2k}$ has a substantial variation from $\tilde{x}_{2k-1}$, which allows for effective contrastive learning.

\section{Manual primary prompts}
\label{sec:primary-prompt}
Table~\ref{table:ManualPrompts_append} shows the primary prompts we used for each task. Those prompts are manually chosen by LM-BFF \cite{gao2021making}.

\setcounter{table}{0}
\renewcommand\thetable{F.\arabic{table}}
\begin{table*}[!ht]
\begin{center}

\begin{tabular}{l c c c c c c}
           \hline
           Task            &     SR    &    RI    &    RS    &  R D   &  EDA    &  ours    \\
           \hline
           SST-2 (acc)     & 90.6 (0.5) & 90.8 (0.4) & \textbf{90.8} (0.4) & 90.8 (0.4) & 90.7 (0.6) &  90.6 (0.1) \\
           Subj (acc)      & 90.4 (1.4) & 90.4 (1.3) & 90.4 (2.5) & 90.3 (1.4) & 90.4 (1.1) &  \textbf{90.4} (1.1) \\
           SST-5 (acc)     & 47.6 (1.4) & 47.0 (1.8) & 46.5 (1.7) & 46.9 (1.9) & 45.2 (2.1) &  \textbf{49.5} (1.1) \\
           CoLA (Matt.)    &  6.0 (5.3) &  6.0 (5.4) &  7.2 (3.8) &  5.6 (3.0) &  5.6 (2.9) &  \textbf{10.2} (5.8) \\
           TREC (acc)      & 80.7 (2.2) & 79.1 (3.8) & 81.2 (2.2) & 82.8 (2.1) & 81.1 (4.3) &  \textbf{83.3} (1.5) \\
           MNLI (acc)      & 60.3 (2.2) & 61.2 (2.1) & 60.7 (2.6) & 60.2 (2.1) & 58.3 (2.5) &  \textbf{64.0} (2.0) \\
           MNLI-mm (acc)   & 62.2 (1.5) & 63.3 (1.3) & 63.0 (1.7) & 62.9 (1.2) & 60.2 (2.1) &  \textbf{65.5} (2.7) \\
           SNLI (acc)      & 63.4 (4.1) & 63.4 (3.8) & 62.3 (3.6) & 62.0 (4.3) & 63.0 (4.2) &  \textbf{69.9} (2.4) \\
           QNLI (acc)      & 63.2 (3.3) & 64.5 (4.6) & 64.0 (4.3) & 64.8 (4.5) & 60.8 (3.6) &  \textbf{66.4} (3.5) \\
           QQP (acc)       & 64.8 (2.8) & 62.9 (2.2) & 62.0 (3.5) & 63.9 (2.1) & 60.4 (5.8) &  \textbf{68.8} (3.8) \\
           RTE (acc)       & 61.2 (3.0) & 62.2 (3.7) & 47.9 (0.8) & 47.9 (0.8) & 64.3 (2.7) &  \textbf{65.1} (3.5) \\
           MRPC (F1)       & 77.0 (3.8) & \textbf{79.2} (4.6) & 78.5 (2.1) & 77.4 (3.2) & 76.2 (6.0) &  78.2 (3.1) \\
           MR (acc)        & 83.3 (1.4) & 85.5 (0.5) & 85.6 (0.6) & 85.2 (0.3) & 85.7 (0.7) &  \textbf{85.8} (0.6) \\
           MPQA (acc)      & 82.6 (2.8) & 82.7 (2.4) & 83.4 (2.4) & 84.3 (2.0) & 83.1 (2.9) &  \textbf{84.6} (1.5) \\
           CR (acc)        & 87.8 (0.9) & 88.1 (0.3) & 87.5 (1.0) & 88.5 (1.0) & 87.9 (0.5) &  \textbf{89.4} (1.0) 
           \\ \hline

\end{tabular}
\caption{Comparing our random templates/demonstrations as data augmentation to synonym replacement (SR), random insertion (RI), random swapping (RS), random deletion (RD) and EDA \cite{wei-zou-2019-eda} (with SR, RI, RS and RD all together) at $20\%$ of input tokens. The results are means of 5 runs with different train-test splits.}
\label{table:CompareEDA20}
\end{center}
\end{table*}

\section{Experiments with RoBERTa-large}
\label{sec:ExpRoBert}
While we use RoBERTa-base to conduct extensive experiments in our main study and ablations, here we compare our framework to LM-BFF using RoBERTa-large. We also include results directly reported from LM-BFF \cite{gao2021making} (henceforth referred to as LM-BFF$\dagger$) , though the comparison between them could be unfair since the results reported in the original LM-BFF paper are:
\begin{itemize}
    \item obtained with an additional sampling strategy to select similar demonstrations (section 6.2 of their paper), which put our results at a disadvantage.
    \item obtained from a set of batch sizes (2,4,8) and learning rates ($1e^{-5}, 2e^{-5}, 5e^{-5}$) and the best models are selected from the validation set. Whereas we show experimental results with models trained with a fixed batch size and a learning rate of $1e^{-5}$.
\end{itemize}
Nevertheless, we show the state-of-the-art results we achieved in Table \ref{table:CompareLMBFF}. We only marginally under-perform LM-BFF$\dagger$ in 2 tasks, possibly due to the reasons listed above.

\setcounter{table}{0}
\renewcommand\thetable{\Alph{section}.\arabic{table}}
\begin{table}[H]
\begin{center}
\small

\begin{tabular}{l c c c c}
           \hline
           Task            & LM-BFF$\dagger$ & LM-BFF$\ddag$ &      ours      \\
           \hline
           SST-2 (acc)     & 92.6 (0.5)       & 91.9 (1.3)      & \textbf{94.2} (0.7) \\
           Subj (acc)      & 92.3 (0.8)       & 91.0 (2.3)      & \textbf{92.4} (0.6) \\
           SST-5 (acc)     & 50.6 (1.4)       & 51.5 (1.4)      & \textbf{54.0} (0.8) \\
           CoLA (Matt.)    & \textbf{18.7} (8.8)     & 11.6 (5.4)       & 18.1 (10.1) \\
           TREC (acc)      & 87.5 (3.2)       & 85.4 (2.6)      & \textbf{89.8} (1.8) \\
           MNLI (acc)      & 70.7 (1.3)       & 70.0 (2.2)      & \textbf{72.4} (2.0) \\
           MNLI-mm (acc)   & 72.0 (1.2)       & 71.8 (2.0)      & \textbf{74.2} (1.9) \\
           SNLI (acc)      & \textbf{79.7} (1.5)       & 79.5 (1.7)      & 79.6 (2.6) \\
           QNLI (acc)      & 69.2 (1.9)       & 67.9 (3.4)      & \textbf{71.1} (6.8) \\
           QQP (acc)       & N/A              & 71.1 (2.3)      & \textbf{74.0} (2.5) \\
           RTE (acc)       & 68.7 (2.3)       & 69.3 (1.1)      & \textbf{71.8} (1.1) \\
           MRPC (F1)       & 77.8 (2.0)       & 77.0 (5.7)      &  77.8 (4.6)\\
           MR (acc)        & 86.6 (2.2)       & 85.6 (3.4)      &  \textbf{89.6} (0.8)\\
           MPQA (acc)      & \textbf{87.0} (1.1)       & 86.9 (1.3)      &  86.9 (1.1)\\
           CR (acc)        & 90.2 (1.2)       & 90.4 (1.2)      & \textbf{91.0} (1.4)
           \\ \hline

\end{tabular}
\caption{Comparison of our method using RoBERT-large to two version of LM-BFF: (1) $\dagger$ the results reported in LM-BFF \cite{gao2021making}. Note that we do \textit{not} have their strategy of sampling similar demonstrations, which may put us at a disadvantage. The training details, hyper-parameters and number of GPUs are also different (2) $\ddag$ The results of LM-BFF using the same batch size, learning rate, training steps, and number of GPUs. These results make the fairest comparison to ours.}
\label{table:CompareLMBFF}
\end{center}
\end{table}

\setcounter{table}{0}
\renewcommand\thetable{\Alph{section}.\arabic{table}}

\section{Augmentation with $20\%$ of input tokens}
\label{sec:Aug}

In the main paper, we compare our augmentation strategy (random templates and random demonstrations) to standard augmentation techniques on 10\% of input tokens for creating multi-view of inputs to apply the SupCon loss. Here, we show additional experimental results with synonym replacement (SR), random insertion (RI), random swapping (RS), random deletion (RD) and EDA \cite{wei-zou-2019-eda} (with SR, RI, RS and RD all together) at $20\%$ of input tokens. Same as before, the model under-performs when using standard augmentations. Results are shown in Table \ref{table:CompareEDA20}.

\setcounter{table}{0}
\renewcommand\thetable{G.\arabic{table}}
\begin{table}[ht]
\begin{center}

\begin{tabular}{l c c}
           \hline
           Task            &  SimCLR    &  SupCon    \\
           \hline
           SST-2 (acc)     & 89.7 (0.8) & \textbf{90.6} (0.1) \\
           Subj (acc)      & 86.4 (1.2) & \textbf{90.4} (1.1) \\
           SST-5 (acc)     & 42.1 (0.8) & \textbf{49.5} (1.1) \\
           CoLA (Matt.)    &  1.8 (3.6) & \textbf{10.2} (5.8) \\
           TREC (acc)      & 60.0 (4.4) & \textbf{83.3} (1.5) \\
           MNLI (acc)      & 52.5 (1.2) & \textbf{64.0} (2.0) \\
           MNLI-mm (acc)   & 53.0 (1.2) & \textbf{65.5} (2.7) \\
           SNLI (acc)      & 60.0 (4.5) & \textbf{69.9} (2.4) \\
           QNLI (acc)      & 53.5 (0.6) & \textbf{66.4} (3.5) \\
           QQP (acc)       & 56.4 (2.0) & \textbf{68.8} (3.8) \\
           RTE (acc)       & 58.6 (2.8) & \textbf{65.1} (3.5) \\
           MRPC (F1)       & 68.2 (6.3) & \textbf{78.2} (3.1) \\
           MR (acc)        & 84.7 (1.1) & \textbf{85.8} (0.6) \\
           MPQA (acc)      & 82.2 (1.6) & \textbf{84.6} (1.5) \\
           CR (acc)        & 88.1 (2.1) & \textbf{89.4} (1.0) 
           \\ \hline

\end{tabular}
\caption{Comparing SimCLR \cite{pmlr-v119-chen20j} and SupCon \cite{SupCon} as different forms of contrastive loss.}
\label{table:CompareSimCLR}
\end{center}
\end{table}

\section{SimCLR vs. SupCon}
\label{sec:SS}

Here, we compare our choice of contrastive loss SupCon \cite{SupCon} to an unsupervised version SimCLR \cite{pmlr-v119-chen20j}. Unsupervised contrastive loss clusters examples at the instance level, \ie, it only pulls the same instance under different views close to each other and push away all the others in a mini-batch. Whereas SupCon clusters examples at the class level.

As shown in Table \ref{table:CompareSimCLR}. SupCon is better than SimCLR in all tasks. In many cases, SimCLR even underperforms the baselines by large margins (see LM-BFF in Table \ref{table:Main}), indicating that learning discriminative features at instance level only can hurt the fine-tuning process.

\setcounter{table}{0}
\renewcommand\thetable{H.\arabic{table}}
\begin{table}[!ht]
\begin{center}
\begin{tabular}{l c c}
          \hline
          Task            &  \texttt{[CLS]}    &  \texttt{[MASK]}    \\
          \hline
          SST-2 (acc)     & 90.4 (0.5) & \textbf{90.6} (0.1) \\
          Subj (acc)      & 90.0 (1.5) & \textbf{90.4} (1.1) \\
          SST-5 (acc)     & 48.2 (0.6) & \textbf{49.5} (1.1) \\
          CoLA (Matt.)    & \textbf{10.7} (5.7) & 10.2 (5.8) \\
          TREC (acc)      & 81.1 (1.4) & \textbf{83.3} (1.5) \\
          MNLI (acc)      & 59.9 (3.7) & \textbf{64.0} (2.0) \\
          MNLI-mm (acc)   & 61.4 (4.0) & \textbf{65.5} (2.7) \\
          SNLI (acc)      & 66.9 (2.5) & \textbf{69.9} (2.4) \\
          QNLI (acc)      & 65.1 (3.1) & \textbf{66.4} (3.5) \\
          QQP (acc)       & 66.0 (3.0) & \textbf{68.8} (3.8) \\
          RTE (acc)       & 63.4 (3.1) & \textbf{65.1} (3.5) \\
          MRPC (F1)       & 77.7 (1.8) & \textbf{78.2} (3.1) \\
          MR (acc)        & 85.5 (0.8) & \textbf{85.8} (0.6) \\
          MPQA (acc)      & 84.2 (1.6) & \textbf{84.6} (1.5) \\
          CR (acc)        & 88.4 (1.2) & \textbf{89.4} (1.0) 
          \\ \hline

\end{tabular}
\caption{Using hidden states at \texttt{[CLS]} tokens or \texttt{[MASK]} tokens as the representations of sentences to perform contrastive learning.}
\label{table:CompareCLS}
\end{center}
\end{table}

\section{ \texttt{[CLS]} vs. \texttt{[MASK]} } \label{sec:CLSorMASK}

SupCon takes the representations of inputs to perform contrastive learning. We use the hidden states at $\texttt{[MASK]}$ tokens as the representations of sentences in the main experiments. Another common choice is to take the hidden states at $\texttt{[CLS]}$ tokens. For example, in standard fine-tuning, the algorithm takes the representation of a sentence at $\texttt{[CLS]}$ token and attaches a linear classifier on top of it.

Based on Table \ref{table:CompareCLS}, applying the contrastive loss at \texttt{[MASK]} tokens is generally better than applying it at \texttt{[CLS]}. This is fairly intuitive, as the final classifications are performed at \texttt{[MASK]} tokens and enforcing class-level discriminative representations explicitly at \texttt{[MASK]} tokens helps models generalize better after fine-tuning.

\section{Adapting ADAPET}
\label{sec:AdA}
For results in Table \ref{table:CompareLoss}, we adapt the open-source code from ADAPET to our codebase. In original ADAPET, there are multiple label words corresponding to a label class (e.g positive class: \textit{"great", "good", "nice"}). To make a fair comparison to LM-BFF and ours, we only apply one label word corresponding to a label class (e.g positive class: \textit{"great"}). The original Label Condition Loss implemented in ADAPET has a hyper-parameter $\alpha$ to control the percentage of input tokens used for masked language modeling (see ADAPET \cite{tam2021improving} for more details). To match the training objective in LM-BFF with only one \texttt{[MASK]} token, we also set the Label Condition Loss to apply to one random token of inputs, i.e., $\alpha=\frac{1}{len(input)}$.

\end{document}